\documentclass[runningheads]{llncs}
\usepackage[T1]{fontenc}
\usepackage{graphicx}
\usepackage{booktabs}
\usepackage[misc]{ifsym}
\newcommand{\corr}{(\Letter)}
\usepackage{mwe}
\usepackage{float}
\usepackage{multirow}
\usepackage{arydshln}
\usepackage{tabularx}
\usepackage{amsmath}
\usepackage{multicol}
\usepackage{dsfont}
\usepackage{hyperref}
\usepackage{comment}
\usepackage{colortbl}

\newcommand\extrafootertext[1]{%
    \bgroup
    \renewcommand\thefootnote{\fnsymbol{footnote}}%
    \renewcommand\thempfootnote{\fnsymbol{mpfootnote}}%
    \footnotetext[0]{#1}%
    \egroup
}

\begin{document}

\title{TAGAL: Tabular Data Generation using Agentic LLM Methods}

\titlerunning{TAGAL: Tabular Data Generation using Agentic LLM Methods}

\author{Benoît Ronval\inst{1} \corr \and
Pierre Dupont\inst{1} \and
Siegfried Nijssen\inst{1,2}}

\authorrunning{B. Ronval et al.}

\institute{UCLouvain, ICTEAM, Belgium \email{\{benoit.ronval,pierre.dupont\}@uclouvain.be}
\and
KU Leuven, DTAI, Belgium \email{siegfried.nijssen@kuleuven.be}
}

\maketitle              

\begin{abstract}

The generation of data is a common approach to improve the performance of machine learning tasks, among which is the training of models for classification. In this paper, we present TAGAL, a collection of methods able to generate synthetic tabular data using an agentic workflow. The methods leverage Large Language Models (LLMs) for an automatic and iterative process that uses feedback to improve the generated data without any further LLM training. The use of LLMs also allows for the addition of external knowledge in the generation process. We evaluate TAGAL across diverse datasets and different aspects of quality for the generated data. We look at the utility of downstream ML models, both by training classifiers on synthetic data only and by combining real and synthetic data. Moreover, we compare the similarities between the real and the generated data. We show that TAGAL is able to perform on par with state-of-the-art approaches that require LLM training and generally outperforms other training-free approaches. These findings highlight the potential of agentic workflow and open new directions for LLM-based data generation methods.\extrafootertext{Accepted at SynDAiTE, ECML PKDD 2025 Workshop}











\end{abstract}


\section{Introduction}



Tabular data is widely used in many domains, from research to industry including healthcare \cite{hernandez2022synthetic}, census data \cite{census}, or finance \cite{sattarov2023findiff} among many others.
However, tabular datasets may present several issues. The data may be imbalanced, with a certain class more represented than the others, data may be difficult to acquire especially in certain domains, e.g. healthcare or finance, or one may not have enough data to represent an entire domain.
A machine learning model learned on such scarce data could perform poorly in a real-world situation. Moreover, certain domains may require that data remain private, making it difficult to share them.
Generated synthetic data may come as a solution for most of these problems. The cost (money, material, or other) to obtain new examples may be largely reduced thanks to generative methods that can sample as many data as desired. Moreover, new examples for underrepresented classes can also be obtained with generative methods, allowing for more balanced datasets. Generative models may also include privacy constraints during the generation \cite{kotal2022privetab}, allowing the sharing of data for a domain with reduced risks on this concern. The process of data generation differs from data augmentation, such as SMOTE \cite{chawla2002smote}, by being able to create entirely new rows instead of doing interpolations between points.

In recent years, Large Language Models (LLMs) have shown to be very efficient for large number of tasks, even tasks for which they were not originally trained for \cite{zhao2023survey}. Thanks to their large number of parameters and attention heads, these models can perform \textit{in-context learning}. Given few examples, called few-shots, an LLM can output similar information to the one provided in these examples.
This in-context learning technique has motivated the use of LLMs with tabular data. LLMs could, for example, be used to classify examples without any training \cite{hegselmann2023tabllm}.
This proves to be an effective approach to detect patterns in data and use them to classify new examples.
Since LLMs can classify tabular examples and are generative by design, a natural extension is to use such models for the generation of tabular data. This may be done with a fine-tuning phase \cite{borisov2022language,zhao2023tabula} or by leveraging the information given by the few-shots directly \cite{fang2025tabgen,kim2025epic}. In this second training-free scenario, LLMs allow for the generation of data based only on a few examples. An additional advantage of LLMs over other generative methods is that they may have background information about the domain of the dataset, potentially leading to more realistic examples.
Even more recently, agentic workflows using LLMs have started to appear.
These methods apply the emergent capabilities of LLMs in automated processes that iteratively improve through the integration of feedback.
Agentic approaches have been shown to improve performance of LLMs for common tasks such as question answering \cite{madaan2023self} but its use for tabular data generation is still limited. We believe that the use of feedback in the generation process may lead to better synthetic data, in terms of utility and quality, while being a training-free approach, reducing the hardware resources needed compared to methods that fine-tune LLMs.
\\

We present \textbf{TAGAL}, for \textbf{Ta}bular data \textbf{G}eneration using \textbf{A}gentic \textbf{L}LMs, composed of three methods: SynthLoop, ReducedLoop, and Prompt-Refine. Each method is based on the same original training-free process and presents different advantages, in terms of computation time or data quality. TAGAL presents the advantages of agentic approaches described before, using one LLM for the generation of the data and one for the feedback.
The use of LLMs in TAGAL allows for interesting features. First, the prompts used in the different parts of the model can be modified to include expert knowledge of a specific domain, direct the generation towards specific values for certain features, or focus the feedback on specific aspects of the generated data. This addition of external knowledge expressed with words can not easily be done with other models that require training.
TAGAL is evaluated on several datasets, looking at both the data utility and quality. TAGAL is further explored by looking at different versions of the prompts and by studying the impact of meta-parameters.

In summary, the main contributions of this work are the following:
\begin{itemize}
    \item We propose TAGAL, a collection of training-free methods using an agentic LLM workflow to generate tabular data, even with limited original data, and leveraging background (from LLMs) or external knowledge to enhance the process.
    TAGAL can generate data with utility and quality on par with methods requiring training while often surpassing other training-free approaches.
    \item We further explore our methods by looking at variations of the prompts and study the impact of the meta-parameters on the generated data.
    \item TAGAL is evaluated against state-of-the-art models using the performance of downstream classification models and metrics to evaluate the similarity between the real and the synthetic data. We also discuss the relationship between the generated data quality and LLM contamination.
\end{itemize}

The implementation of TAGAL is available in the following repository: \url{https://github.com/bronval/TAGAL-Tabular-Data-Generation-with-LLMs/}


\section{Related Works}
\label{sec:related works}

\subsection{Generative Models with Training}
\label{sec:model_train}

Different deep models have been presented to generate synthetic tabular data. CTGAN \cite{xu2019modeling} adapts the GAN architecture by introducing a mode-specific normalization to model the continuous features. It is also designed for conditional generation, meaning some values can be forced for the generated examples. Another adaptation is TabDDPM \cite{kotelnikov2023tabddpm} which is based on a diffusion process. It separates between two types: a multinomial diffusion for the categorical and binary features, and a Gaussian diffusion for the numerical features. The first type uses uniform noise over the different classes in the example while the second one uses Gaussian distributions in the noising and denoising parts. The use and fine-tuning of LLMs have also been explored for this goal. GReaT \cite{borisov2022language} first transforms the tabular data into text, including a shuffle of the feature orders to allow for conditional generation. An LLM like GPT-2 (or distil-GPT-2) is then fine-tuned on this text data. This simple approach proves to be very effective. 
Tabula \cite{zhao2023tabula} proposes an updated version of GReaT by using padding tokens to align the features at the token level.
Both GReaT and Tabula fine-tune LLMs by maximizing the next token prediction, as done in the original training of LLMs.

\subsection{Generative Models without LLM Training}
\label{sec:model_notrain}

Previously introduced generative models all require training on the original examples. On the other hand, foundation models like LLMs, already trained on large amount of data, can be used directly for many tasks, performing well without the need of fine-tuning \cite{zhao2023survey}. The large number of parameters and enormous amount of training data allow for emergent capabilities, such as in-context learning described before.
This capability, along with pattern detection, is at the core of EPIC \cite{kim2025epic}. This method designs prompts only made of few-shot original examples grouped by classes. No other prompt or information is given to the model. EPIC serializes the data using a CSV format where the first line contains the names of the features and each following line is an example with values separated by commas.
To reach the expected amount of synthetic data, EPIC reuses the prompt with different few-shot examples until the amount is met.
MALLM-GAN \cite{ling2024mallm} is another LLM method, structured with an optimization loop. It leverages two LLMs: one for data generation, and one to optimize the prompt used for the data generation. 
This second part uses a logistic regression to distinguish real examples from generated ones. The obtained detection score is then given to the LLM of this second part, which is prompted to use this score to improve the prompt used to generate data.
A more recent approach is TabGen-ICL \cite{fang2025tabgen}, which focuses on the selection of examples to be put in the prompt of the LLM tasked to generate tabular data. This method compares the distributions of real and generated data and select the next few-shots examples among the real examples that fall outside of the generated distribution.

\subsection{Data Contamination}
Although LLMs can serve as classifiers \cite{hegselmann2023tabllm}, there has been some criticism about this use. Given the huge size of their training set, LLMs likely have seen popular test data during training \cite{ronval2025detection}.
This may result in an optimistically biased "test" performance. Yet, data contamination may allow LLMs to use background knowledge from datasets similar to the one for which it generates data.


\subsection{Agentic LLM}

The term agentic LLM refers to an LLM-based workflow where the model automatically iterates on its outputs, leveraging feedback mechanisms to improve its results. It may include the use of external tools, such as predefined functions or other LLMs. Unlike traditional prompt-response use, an agentic LLM leverages structured reasoning, using for example chain-of-thought \cite{wei2022chain}, over several iterations to self-correct and reach a defined goal.
Early work in this direction like Self-Refine \cite{madaan2023self} shows that making the LLM iterate over its answers can improve the performance of such models on question-answering tasks.
Agentic LLMs may include the concept of multi-agent LLMs as well \cite{talebirad2023multi}. In this setup, each agent may have a different role to reach a common goal. This split in different agents allows each of them to be specialized in its part of the process. MALLM-GAN, described above, may be seen as an agentic approach to generate tabular data.

In this work, the focus is on a multi-agent iterative approach that leverages the in-context learning capability of LLMs, without any limitation on the feedback, to generate synthetic tabular data.


\section{TAGAL: Tabular Data Generation using Agentic LLM}
\label{sec:method}

TAGAL is a collection of three agentic LLM-based methods: SynthLoop, ReducedLoop, and Prompt-Refine. It is training-free and model agnostic such that any LLM, closed or open-source, may be used. Furthermore, the use of LLMs and the absence of training allow the methods from TAGAL to work with limited amount of data thanks to the large number of parameters of these models that allows for in-context learning with few-shot examples.

\subsection{SynthLoop}
\label{sec:synthloop}

\begin{figure}[t]
    \centering
    \includegraphics[width=\linewidth]{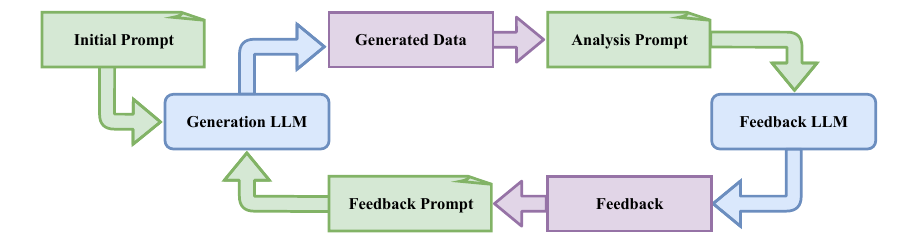}
    \caption{Overview of the SynthLoop method to generate tabular data. SynthLoop includes two LLMs, one generating the new examples and one providing feedback.}
    \label{fig:diagram_loop}
\end{figure}

The first model of TAGAL, SynthLoop, presents the basic version of the approach. It is displayed in figure \ref{fig:diagram_loop}.
It starts with the design of a system prompt to describe the task. This prompt includes guidelines on the creation of synthetic examples ("follow the distribution of the original dataset", "find patterns in the original examples", ...) and prompts the model to think step by step. It incorporates information about the features of the dataset as well: the name, the type (int, float, ...), and whether it is categorical or numerical. For categorical features, the possible values are listed with their distribution in the original dataset. For numerical features, it gives the mean, median, and standard deviation of the values.
Next, a user prompt with few-shot examples asks the \textbf{generation LLM} to generate tabular examples. Following the conclusions from EPIC \cite{kim2025epic}, the data is serialized with a CSV style format and groups the few-shot examples by classes. These first two prompts, system and user, are referred to as the \textbf{initial prompt}.

The \textbf{analysis prompt} is composed of a system and user prompts. The system prompt asks the \textbf{feedback LLM} to proceed step by step to criticize the generated data by looking at their strengths and weaknesses and finally give recommendations to improve the data. To do this, the same information about the dataset are placed in this prompt. Even if both generation and feedback LLMs use the same information, the important aspect is obtaining feedback to correct the generation LLM, as described in Self-Refine \cite{madaan2023self}.
The analysis user prompt then asks the feedback LLM to generate feedback for the generated data. Here, our method differs from MALLM-GAN \cite{ling2024mallm} by using an LLM for the whole feedback process, both at the critic and prompt optimization steps, instead of using a single external score to optimize the prompt used for data generation.

The obtained feedback is placed in the next user prompt, which asks the generation LLM to generate new examples following the additional recommendations from the feedback. This ends the first iteration of the feedback loop. The iterations are repeated a few times, decided by the user to control the total generation time. The final generated data outputted by the model are the examples obtained during the last loop. The completion of all the iterations is referred to as a run.
During the whole process, a distinction is made between the generation and feedback conversation histories. For the generation, it contains the initial prompt, the generated examples, and the user prompt with the feedback. The conversation history grows with the iterations which means it contains $[initial \, prompt, gen. \, examples, feedback, gen. \, examples, feedback, ... ]$ until it reaches the end of the process. The feedback history follows the same idea. It contains $[initial \, analysis \, prompt, feedback, gen. \, examples, feedback, ...]$ and grows until the end of the process. Each conversation history is passed to the corresponding LLM each time it is called for its task. This distinction between the conversation histories is why we refer to two different LLMs, even if they may be the same base models (Llama 3.1 8B by default).

A single run, i.e. the completion of all (e.g. 3) iterations, may not produce the required amount of synthetic data directly due to the context size of LLMs. To remedy this, SynthLoop restarts the whole process by resetting both conversation histories and sampling other few-shot examples for the initial prompt to improve the diversity of generated data. This allows the process to continuously use feedback during the generation while using different few-shot examples for more diversity. However, SynthLoop is time consuming as it needs to perform several runs if one requires thousands of synthetic examples.

\subsection{ReducedLoop}

The ReducedLoop version is split into two phases. The first phase follows the exact same iterative process as described in the SynthLoop version. At the end of the first run, it starts the second phase. To meet the desired amount of synthetic data, the exact same conversation history of the generation part obtained after the run is resubmitted.
This varies from SynthLoop as it does not restart the full iteration process by resetting the conversation histories but instead considers that one run is enough.
The major advantage of this method is the reduced time necessary to reach the required amount of synthetic data while keeping access to the full conversation history. It allows the generation LLM to still leverage information gained from the iterative process, including the feedback given by the feedback LLM.
However, one major disadvantage is that ReducedLoop may output more duplicates in the generated data due to same conversation history being submitted multiple times to get the synthetic tabular data.

\begin{figure}[t]
    \centering
    \includegraphics[width=\linewidth]{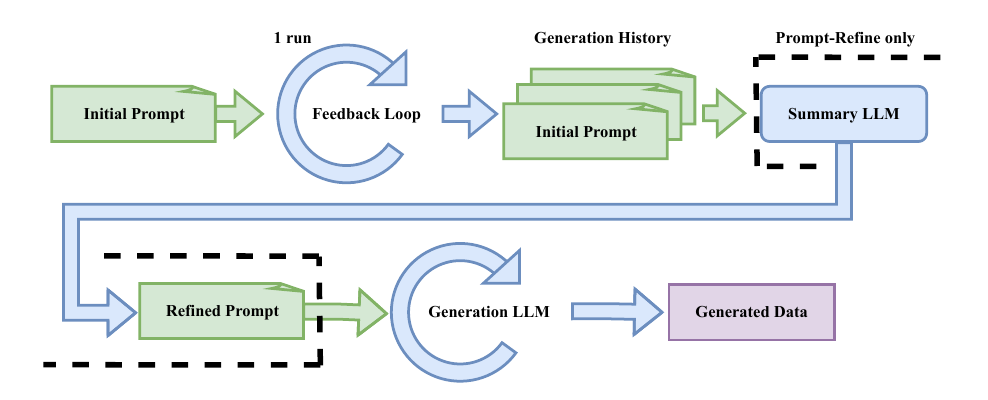}
    \caption{Overview of the ReducedLoop and Prompt-Refine approaches. The \textbf{Feedback Loop} part is the iterative process shown in figure \ref{fig:diagram_loop}. Parts between dotted lines are only executed for Prompt-Refine.}
    \label{fig:promptrefine}
\end{figure}

\subsection{Prompt-Refine}

\noindent Prompt-Refine, illustrated in figure \ref{fig:promptrefine}, is the third and last method from TAGAL. It splits into three phases. The first is common with SynthLoop and ReducedLoop, where one run of (e.g. 3) iterations is performed and both conversation histories are completed with the prompts and outputs. The second phase uses a third LLM, called \textbf{summary LLM}, to create a new prompt. This LLM receives the entire generation's conversation history and is tasked to output a prompt summarizing the important information which will be used to generate other tabular data. We let the LLM creates this prompt automatically, continuing the concept of agentic LLM.
The only element manually included in this prompt are few-shot examples to ensure the generation LLM follows the template of the data when generating new ones.
We refer to this final prompt as the \textbf{refined prompt}.
The idea is that the summary LLM will find information from the conversation history that may be important or repeated across the feedbacks but that a human user could not think about when creating a prompt to generate data. Moreover, the reduction of the conversation history to a single prompt allows for a cost reduction, both monetary when using paid APIs and computational as the LLM has less input tokens to process.
The third phase is the repeated use of the refined prompt to generate tabular examples until the required amount of data is reached. At each call of the generation LLM, this prompt has no other modification than the sampling of different few-shot examples.
Prompt-Refine can be seen as the same approach as ReducedLoop where the conversation history is summarized to a single prompt by an LLM.

Compared to the other methods, Prompt-Refine allows for a higher gain of time and cost due to the lower number of tokens to process. Moreover, the use of few-shot examples allows us to vary the examples shown to the generation LLM, which could lead to more diverse generated examples. However, we note that the quality of the generated examples largely depends on the quality of the refined prompt generated by the feedback LLM.

\subsection{Feedback and Refined Prompt Example}

\begin{table}[htbp]
    \centering
    \caption{Example of information from the feedbacks automatically added in the refined prompt for the Adult dataset using DeepSeek-v3.}
    \begin{tabular}{|p{1.0\linewidth}|}
    \hline
\textbf{Feedback Incorporation}:\\

   - \textbf{Diversity in `race`}: Include more examples of underrepresented categories like "Black," "Asian-Pac-Islander," and "Amer-Indian-Eskimo."\\
   
   - \textbf{Diversity in `native.country`}: Add more examples from underrepresented countries like "Mexico," "Germany," or "India."\\
   
   - \textbf{Variability in `capital.gain` and `capital.loss`}: Introduce higher non-zero values and a wider range of losses.\\
   
   - \textbf{Range of `hours.per.week`}: Include more extreme values (e.g., <20 or >60) to better match the original dataset's distribution.\\
   
   - \textbf{Income Distribution}: Adjust the `income` distribution to better reflect the original dataset's proportions (75.82\% <=50K and 24.18\% >50K).\\
   
   - \textbf{Unique Entries}: Ensure all generated examples are unique and free from duplicates.
   
   - \textbf{Less Common Categories}: Include more examples of less common categories in `education` (e.g., "Preschool" or "Doctorate"), `marital.status` (e.g., "Married-AF-spouse" or "Married-spouse-absent"), and `occupation` (e.g., "Armed-Forces" or "Priv-house-serv").\\
    \hline
    \end{tabular}
    
    \label{tab:refined_example}
\end{table}

The idea of the Prompt-Refine approach is to use an LLM to summarize the conversation history with the feedback to include elements that one could not have thought of or that could be specific to the LLM, such as missing information in its background knowledge. Table \ref{tab:refined_example} displays an example of this idea.
This automatically generated refined prompt contains the description of the task and the information of the dataset (not displayed here for space reasons), but also some elements from the automatically generated feedback, highlighting and trying to correct flaws observed during the feedback loop.


\section{Experimental Setup}

\subsection{Datasets}

TAGAL is evaluated on four tabular datasets, detailed in Table \ref{tab:datasets}. We highlight that the Thyroid dataset was released after the cutoff of most current LLMs and is therefore unlikely to appear in their training sets. This allows an evaluation of the methods without data contamination and may show the advantage of external knowledge (here, the information about the features) added to the prompts. Each dataset is split into 80\% training and 20\% test sets before the generation part to limit that no test data may influence the generation, which could result in the contamination of the downstream classifier model. This only limits the contamination since LLMs may have seen some test data during their training.

Although the multi-class setting is not studied in this work, the generation methods of TAGAL consider the target feature as any other feature from the given dataset. Therefore, TAGAL can be used with multi-class datasets without modification of the methods.

\begin{table}[t]
    \centering
    \caption{Datasets details.}
    \resizebox{\textwidth}{!}{
    \begin{tabular}{lcccr}
    \hline
         \textbf{Name} & \textbf{\# Examples} & \textbf{\# Features} & \textbf{Classes} & \textbf{Target name } \\
         \hline
         \hline
         Adult & 35,561 & 15 & <=50K (75.82\%) / >50K (24.18\%) & Income\\
         Bank & 45,211 & 17 & no (88.22\%) / yes (11.68\%) & y\\
         German & 1000 & 10 & good (70\%) / bad (30\%) & Risk\\
         Thyroid & 383 & 17 & No (73.2\%) / Yes (26.8\%) & Recurred\\
         \hline
    \end{tabular}
    }
    \label{tab:datasets}
\end{table}

\subsection{Evaluation Metrics}

The synthetic data evaluation considers two aspects: 1) its utility in a downstream classification task and 2) the similarity between real and synthetic examples.
For the first aspect, \textbf{machine learning utility} is assessed in two setups: \textit{Train on Synthetic, Test on Real} (TSTR) where the downstream model is learned only on generated data, and \textit{combined}, where the train set is made of 50\% real data and 50\% generated data. Three classifiers --- Random Forest, LightGBM, and XGBoost --- are evaluated over 5 independent training runs to smooth randomness. The averaged test set performance is reported.
The second aspect is evaluated with \textbf{precision} and \textbf{recall} \cite{sajjadi2018assessing} between the real and synthetic examples. Formal definitions are given by \cite{naeem2020reliable}:



\begin{multicols}{2}
\begin{equation}
    \text{precision} := \frac{1}{M}\sum_{j=1}^{M} 1_{Y_j \in \text{manifold}(X_1, ..., X_N)}
    \nonumber
\end{equation}\break
\begin{equation}
    \text{recall} := \frac{1}{N} \sum_{i=1}^{N} 1_{X_i \in \text{manifold}(Y_1, ..., Y_M)}
    \nonumber
\end{equation}
\end{multicols}

\noindent where $M$ and $N$ are the number of synthetic and real examples. A manifold is represented by $\text{manifold}(X_1, ..., X_N) := \bigcup_{i=1}^{N} B(X_i, NND_k(X_i))$ where $B(x, r)$ is the sphere in $\mathds{R}^D$ centered around $x$ with radius $r$ in the space of $D$ features. $NND_k(X_i)$ is the distance between $X_i$ and the $k^{th}$ closest point. Following \cite{qian2023synthcity}, we use $k=5$ in the experiments.

Additionally, the number of collisions in the generated data is monitored. A collision is a synthetic data being identical to a real one, missing the goal of the generation process. The collisions are removed when evaluating TSTR, precision, and recall as they do not respect the evaluation setup for TSTR and may wrongly influence the similarities for precision and recall.

\subsection{Baseline Statistical Generation}
\label{sec:statgen}

In addition to the models from Section \ref{sec:related works}, we implement a simple generative baseline. Given a dataset, it generates values for the categorical features by randomly sampling one of the possible values based on the distribution in the feature. For the numerical features, it samples from a Gaussian distribution based on the mean and standard deviation from the feature.

\subsection{Meta-Parameters}

We detail here the different meta-parameters used in TAGAL, including the meta-parameters for the LLMs.
The maximal number of iterations of the loop is fixed at 3. The prompt contains 20 few-shot examples for each class in the initial prompt and asks for 2500 examples to be generated.
Concerning the LLMs, we use Llama3.1 8B \cite{dubey2024llama} for our approach and EPIC.
For GReaT and Tabula, we use a distil-gpt2 model \cite{radford2019language} that is fine-tuned according to these methods. We also test TAGAL with larger models, namely GPT4o \cite{hurst2024gpt} and DeepSeek-v3 \cite{liu2024deepseek}.
Except when mentioned otherwise, we consistently use a temperature of 0.7 for all LLMs in TAGAL.
We fix the number of maximum generated tokens to 16,384 for the data generation and 2,048 for the feedback.



\section{Results}


\subsection{Comparison with Other Models}

\begin{table}[htbp]
    \centering
    \caption{
    Comparison of TAGAL methods (3 last rows) using Llama 3.1 8B, with models from the literature. Numbers beside the dataset name indicate the size of training sets for the downstream utilities, which may be smaller than the 20\% test set due to the removal of collisions when evaluating TSTR, precision, and recall. Original denotes test set classification score when trained on original data. ROC AUC score is used for the utilities and average performance across all runs of the classifiers is reported. $\uparrow$ means the higher the better and best scores are indicated in bold.}
    

\resizebox{\textwidth}{!}{
    
\begin{tabular}{clccccc}
\hline
\textbf{Dataset} & \textbf{Model} & \textbf{U. TSTR} $\uparrow$ & \textbf{U. Comb} $\uparrow$ & \textbf{Precision} $\uparrow$ & \textbf{Recall} $\uparrow$ & \textbf{Collisions} [\%] \\
\hline
\hline
\multirow{10}{*}{\rotatebox[origin=c]{90}{Adult (1108)}} & Original & \textbf{0.89} & - & - & - & - \\
& Stat. Gen. & 0.53 & 0.87 & 0.60 & 0.89 & 0.00 \\
& CTGAN & 0.86 & 0.88 & 0.76 & 0.89 & 0.00 \\
& TabDDPM & 0.87 & 0.89 & 0.92 & \textbf{0.90} & 0.00 \\
& GReaT & 0.87 & \textbf{0.89} & \textbf{0.99} & 0.69 & 0.00 \\
& Tabula & 0.88 & 0.89 & 0.99 & 0.77 & 0.40 \\
& EPIC & 0.53 & 0.88 & 0.91 & 0.37 & 56.51 \\
& SynthLoop & 0.74 & 0.88 & 0.64 & 0.71 & 12.28 \\
& ReducedLoop & 0.75 & 0.88 & 0.84 & 0.53 & 44.08 \\
& Prompt-Refine & 0.85 & 0.88 & 0.94 & 0.86 & 3.16 \\

\hline
\multirow{10}{*}{\rotatebox[origin=c]{90}{Bank (941)}} & Original & \textbf{0.89} & - & - & - & - \\
& Stat. Gen. & 0.61 & 0.85 & 0.80 & 0.93 & 0.00 \\
& CTGAN & 0.69 & 0.85 & 0.88 & 0.91 & 0.00 \\
& TabDDPM & 0.86 & 0.88 & 0.93 & \textbf{0.94} & 0.00 \\
& GReaT & 0.85 & 0.87 & 0.99 & 0.85 & 0.00 \\
& Tabula & 0.86 & 0.88 & 0.98 & 0.86 & 0.00 \\
& EPIC & 0.66 & \textbf{0.90} & 0.90 & 0.52 & 64.10 \\
& SynthLoop & 0.70 & 0.87 & 0.54 & 0.55 & 0.08 \\
& ReducedLoop & 0.49 & 0.87 & 0.20 & 0.43 & 32.52 \\
& Prompt-Refine & 0.59 & 0.87 & \textbf{0.99} & 0.83 & 19.08 \\

\hline
\multirow{10}{*}{\rotatebox[origin=c]{90}{German (800)}} & Original & \textbf{0.78} & - & - & - & - \\
& Stat. Gen. & 0.51 & 0.75 & 0.93 & 0.66 & 0.00 \\
& CTGAN & 0.46 & 0.67 & 0.92 & 0.62 & 45.84 \\
& TabDDPM & 0.68 & 0.71 & 0.95 & 0.64 & 48.48 \\
& GReaT & 0.55 & 0.69 & \textbf{0.98} & 0.58 & 47.60 \\
& Tabula & 0.59 & 0.57 & 0.98 & \textbf{0.96} & 67.44 \\
& EPIC & 0.45 & 0.71 & 0.92 & 0.21 & 45.58 \\
& SynthLoop & 0.58 & \textbf{0.76} & 0.71 & 0.50 & 0.24 \\
& ReducedLoop & 0.49 & 0.74 & 0.46 & 0.57 & 0.00 \\
& Prompt-Refine & 0.49 & 0.70 & 0.68 & 0.41 & 0.00 \\

\hline
\multirow{10}{*}{\rotatebox[origin=c]{90}{Thyroid (306)}} & Original & 0.97 & - & - & - & - \\
& Stat. Gen. & 0.47 & 0.91 & 0.51 & 0.86 & 0.56 \\
& CTGAN & 0.56 & 0.94 & 0.56 & 0.85 & 0.12 \\
& TabDDPM & 0.96 & 0.98 & 0.94 & \textbf{0.97} & 4.72 \\
& GReaT & 0.92 & 0.97 & 0.75 & 0.71 & 8.28 \\
& Tabula & 0.98 & 0.98 & 0.79 & 0.87 & 23.97 \\
& EPIC & 0.98 & \textbf{0.98} & \textbf{1.00} & 0.55 & 73.51 \\
& SynthLoop & 0.95 & 0.98 & 0.68 & 0.78 & 0.44 \\
& ReducedLoop & 0.91 & 0.98 & 0.91 & 0.30 & 0.32 \\
& Prompt-Refine & \textbf{0.98} & 0.97 & 0.93 & 0.85 & 7.08 \\

\hline
\end{tabular}
}
    \label{tab:competitors}
\end{table}

\begin{table}[t]
    \centering
    \caption{Comparison of TAGAL methods using different LLMs. Test set ROC AUC score is used for utilities in the same setup as in Table \ref{tab:competitors}.}
\resizebox{\textwidth}{!}{    
\begin{tabular}{cllccccc}
\hline
\textbf{Dataset} & \textbf{LLM} & \textbf{Model} & \textbf{U. TSTR} $\uparrow$ & \textbf{U. Comb} $\uparrow$ & \textbf{Precision} $\uparrow$ & \textbf{Recall} $\uparrow$ & \textbf{Collisions} [\%] \\
\hline
\hline
\multirow{10}{*}{\rotatebox[origin=c]{90}{Adult (1398)}} & & Original & \textbf{0.90} & - & - & - & - \\
\cdashline{2-8}
& \multirow{3}{*}{Llama 3.1} & SynthLoop & 0.79 & 0.88 & 0.64 & 0.71 & 12.28 \\
& & ReducedLoop & 0.75 & 0.88 & 0.84 & 0.53 & 44.08 \\
& & Prompt-Refine & 0.85 & 0.89 & \textbf{0.94} & 0.86 & 3.16 \\
\cdashline{2-8}
& \multirow{3}{*}{GPT-4o} & SynthLoop & 0.88 & 0.89 & 0.73 & \textbf{0.93} & 0.00 \\
& & ReducedLoop & 0.88 & 0.89 & 0.78 & 0.88 & 0.00 \\
& & Prompt-Refine & 0.87 & 0.89 & 0.88 & 0.86 & 4.80 \\
\cdashline{2-8}
& \multirow{3}{*}{DeepSeek-v3} & SynthLoop & 0.88 & \textbf{0.89} & 0.90 & 0.88 & 8.44 \\
& & ReducedLoop & 0.80 & 0.88 & 0.92 & 0.45 & 0.00 \\
& & Prompt-Refine & 0.87 & 0.89 & 0.93 & 0.76 & 3.24 \\
\hline
\multirow{10}{*}{\rotatebox[origin=c]{90}{Bank (981)}} & & Original & \textbf{0.88} & - & - & - & - \\
\cdashline{2-8}
& \multirow{3}{*}{Llama 3.1} & SynthLoop & 0.71 & 0.87 & 0.54 & 0.55 & 0.08 \\
& & ReducedLoop & 0.49 & 0.87 & 0.20 & 0.43 & 32.52 \\
& & Prompt-Refine & 0.58 & 0.87 & \textbf{0.99} & 0.83 & 19.08 \\
\cdashline{2-8}
& \multirow{3}{*}{GPT-4o} & SynthLoop & 0.82 & 0.87 & 0.84 & 0.94 & 0.00 \\
& & ReducedLoop & 0.61 & 0.85 & 0.88 & 0.73 & 0.00 \\
& & Prompt-Refine & 0.86 & \textbf{0.90} & 0.90 & \textbf{0.95} & 60.76 \\
\cdashline{2-8}
& \multirow{3}{*}{DeepSeek-v3} & SynthLoop & 0.85 & 0.88 & 0.92 & 0.90 & 7.12 \\
& & ReducedLoop & 0.71 & 0.86 & 0.95 & 0.06 & 0.00 \\
& & Prompt-Refine & 0.84 & 0.87 & 0.95 & 0.83 & 1.44 \\
\hline
\multirow{10}{*}{\rotatebox[origin=c]{90}{Thyroid (306)}} & & Original & 0.97 & - & - & - & - \\
\cdashline{2-8}
& \multirow{3}{*}{Llama 3.1} & SynthLoop & 0.95 & 0.98 & 0.68 & 0.78 & 0.44 \\
& & ReducedLoop & 0.91 & 0.98 & 0.91 & 0.30 & 0.32 \\
& & Prompt-Refine & \textbf{0.98} & 0.97 & 0.93 & 0.85 & 7.08 \\
\cdashline{2-8}
& \multirow{3}{*}{GPT-4o} & SynthLoop & 0.97 & 0.97 & 0.70 & 0.95 & 2.80 \\
& & ReducedLoop & 0.97 & \textbf{0.99} & 0.69 & 0.84 & 5.52 \\
& & Prompt-Refine & 0.98 & 0.97 & 0.80 & 0.94 & 63.44 \\
\cdashline{2-8}
& \multirow{3}{*}{DeepSeek-v3} & SynthLoop & 0.97 & 0.97 & 0.88 & 0.92 & 42.44 \\
& & ReducedLoop & 0.96 & 0.97 & \textbf{0.98} & 0.45 & 19.28 \\
& & Prompt-Refine & 0.97 & 0.98 & 0.97 & \textbf{0.97} & 10.60 \\
\hline
\end{tabular}
    }
    \label{tab:llms}
\end{table}






Table \ref{tab:competitors} reports the results of the 3 TAGAL methods compared to the existing techniques presented in sections \ref{sec:model_train}, \ref{sec:model_notrain}, and \ref{sec:statgen}.
Looking at the TSTR utility (U. TSTR), training classifiers only on original data proves to be better than training on synthetic data. This is expected as the original training data are the most similar to the test set, made exclusively of real data. On most datasets, at least one of the TAGAL methods can reach performance close to the ones obtained by trained generative models. Moreover, the best method from TAGAL always outperforms EPIC, the other considered LLM training-free approach. These results highlight that the data generated by TAGAL may act as a replacement for the real data.
Some datasets, like German, seem to make the generation task more difficult for this metric for most methods. This may show that certain datasets have an inherent difficulty for both the classification and the generation tasks.

If real and synthetic data are mixed, results for the utility (U. Comb) tend to increase compared to TSTR, closing in and sometimes improving the original classification performance. TAGAL methods perform well, either by scoring the best utility or by being on par with the other methods.
In this particular setup, the performance for EPIC drastically increases for most datasets compared to the TSTR case. We argue that this is due to the very high number of collisions between real and synthetic data produced by this model, which results in a majority of data augmentation instead of data generation.
We highlight that TAGAL methods have very high utility, even improving the score of the original data, on the Thyroid dataset which has a limited amount of data and that is unlikely to have been seen during the pre-training of the LLMs. This proves that the methods from TAGAL can use the information from the few-shot examples and do not need a large amount of data to work well.

Concerning precision and recall, generative models with training perform generally better than the LLM training-free approaches, with some exceptions. We argue this result is expected since these models learn the distributions of the real data, thus generating examples closer to the real data. On the other hand, LLM training-free approaches perform well on some datasets but may benefit from more few-shot examples to better detect the patterns in the real data.

Collisions may happen for all models, especially on smaller datasets due to the limited amount of data, but more frequently with the LLM training-free approach. We explain this by the few-shot examples given to the LLMs, which may push them to copy these given examples by altering the output probabilities of the model. This is even more the case in EPIC, where no other indication than the few-shot examples is provided. TAGAL methods are also based on few-shot examples but the feedback process may drastically reduce the number of collisions compared to EPIC, making TAGAL models output more "truly generated" examples.
Among the TAGAL methods, ReducedLoop seems to output more collisions on average across the different datasets. This is explained by the repeated use of the conversation history, which makes the diversity of the generated examples rely mostly on the temperature parameter.

The results for the Thyroid dataset show that, even though the LLMs have arguably not been trained on this dataset, the information in the prompts and the patterns shown in the few-shot examples can be enough to obtain good synthetic data. This opens the use of TAGAL methods on future datasets.
\\




Table \ref{tab:llms} shows the use of different LLMs in the feedback loop of TAGAL. The basic setup using Llama 3.1 (8B parameters) is compared to GPT4o and DeepSeek-v3 (671B parameters).
Larger models tend to improve the results, approaching the performance obtained with original data. We hypothesize that this result can be explained by the larger number of parameters present in such models.
We believe that such models are more prone to follow feedback and thus obtain examples of higher quality thanks to these higher capabilities.
However, we highlight that the Llama model still performs well and sometimes obtains better scores than larger models, despite being significantly smaller in terms of parameters. This shows that TAGAL can be used with such open-source models that do not require large hardware to execute.

\begin{table}[t]
    \centering
    \caption{Modification of selected meta-parameters for SynthLoop and PromptRefine (using Llama 3.1 8B).}
    \resizebox{\textwidth}{!}{  
\begin{tabular}{cllccccc}
\hline
\textbf{Dataset} & \textbf{Model} & \textbf{Variant} & \textbf{U. TSTR} $\uparrow$ & \textbf{U. Comb} $\uparrow$ & \textbf{Precision} $\uparrow$ & \textbf{Recall} $\uparrow$ & \textbf{Collisions} [\%] \\
\hline
\hline
\multirow{15}{*}{\rotatebox[origin=c]{90}{Adult (1797)}} & & Original & \textbf{0.90} & - & - & - & - \\
\cdashline{2-8}
& \multirow{6}{*}{SynthLoop} & baseline & 0.79 & 0.88 & 0.64 & 0.71 & 12.28 \\
& & temp. 0.9 & 0.84 & 0.89 & 0.86 & 0.85 & 4.40 \\
& & 30 shots & 0.77 & 0.88 & 0.93 & 0.80 & 28.12 \\
& & cat first & 0.71 & 0.88 & 0.34 & 0.76 & 0.48 \\
& & num first & 0.75 & 0.87 & 0.78 & 0.70 & 2.00 \\
& & fshots feedback & 0.81 & 0.88 & 0.62 & \textbf{0.89} & 19.92 \\
\cdashline{2-8}
& \multirow{6}{*}{Prompt-Refine} & baseline & 0.86 & \textbf{0.89} & 0.94 & 0.86 & 3.16 \\
& & temp. 0.9 & 0.79 & 0.88 & 0.84 & 0.85 & 0.08 \\
& & 30 shots & 0.80 & 0.87 & \textbf{0.96} & 0.65 & 7.76 \\
& & cat first & 0.64 & 0.87 & 0.81 & 0.66 & 0.36 \\
& & num first & 0.73 & 0.87 & 0.88 & 0.43 & 1.48 \\
& & fshots feedback & 0.83 & 0.89 & 0.85 & 0.55 & 7.20 \\
\hline
\multirow{15}{*}{\rotatebox[origin=c]{90}{Thyroid (306)}} & & Original & 0.97 & - & - & - & - \\
\cdashline{2-8}
& \multirow{6}{*}{SynthLoop} & baseline & 0.95 & 0.98 & 0.68 & 0.78 & 0.44 \\
& & temp. 0.9 & 0.95 & 0.97 & 0.59 & 0.81 & 0.96 \\
& & 30 shots & 0.93 & 0.97 & 0.66 & 0.59 & 0.32 \\
& & cat first & 0.91 & 0.97 & 0.46 & 0.53 & 1.04 \\
& & num first & 0.97 & 0.96 & 0.80 & 0.66 & 0.56 \\
& & fshots feedback & 0.98 & 0.98 & 0.82 & 0.67 & 0.80 \\
\cdashline{2-8}
& \multirow{6}{*}{Prompt-Refine} & baseline & \textbf{0.98} & 0.97 & 0.93 & \textbf{0.85} & 7.08 \\
& & temp. 0.9 & 0.88 & 0.98 & 0.58 & 0.07 & 0.24 \\
& & 30 shots & 0.93 & 0.98 & 0.78 & 0.68 & 0.56 \\
& & cat first & 0.91 & 0.94 & 0.78 & 0.78 & 4.16 \\
& & num first & 0.70 & \textbf{0.98} & 0.93 & 0.61 & 1.64 \\
& & fshots feedback & 0.95 & 0.97 & \textbf{0.96} & 0.66 & 2.80 \\
\hline
\end{tabular}
}
    \label{tab:variants}
\end{table}

\subsection{Study of Meta-Parameters}

We further study the TAGAL methods by changing selected meta-parameters to identify potential improvements.
Different variants are considered: \textbf{temp 0.9}, where the temperature for the generation LLM is set to 0.9 instead of 0.7; \textbf{30 shots}, using 30 few-shot examples for each class instead of 20; \textbf{cat first}, where the features have been ordered with the categorical columns appearing first; \textbf{num first}, the same for numerical columns; \textbf{fshots feedback}, where the same few-shot examples used for the generation LLM are passed to the feedback LLM.

Table \ref{tab:variants} shows the results for this experiment. The baseline approach, presented in previous tables, seems to perform well compared to the variants, especially for the utilities. Even if some variants may sometimes outperform the others, no clear winner seems to emerge from this comparison. We highlight that using a higher temperature for the generation LLM often leads to less collisions and duplicates (the same data being generated multiple times), as the generation is more diverse, but could lead to major collapse of the quality metrics, such as the recall for the Thyroid dataset using Prompt-Refine. We notice that giving the few-shots to the feedback LLM as well may in some cases help to direct the generation towards data that more closely matches the real distribution. 

\begin{table}[t]
    \centering
    \caption{Comparison of prompt variants for SynthLoop and Prompt-Refine (using Llama 3.1 8B).}

\resizebox{\textwidth}{!}{   
\begin{tabular}{cllccccc}
\hline
\textbf{Dataset} & \textbf{Model} & \textbf{Variant} & \textbf{U. TSTR} $\uparrow$ & \textbf{U. Comb} $\uparrow$ & \textbf{Precision} $\uparrow$ & \textbf{Recall} $\uparrow$ & \textbf{Collisions} [\%] \\
\hline
\hline
\multirow{9}{*}{\rotatebox[origin=c]{90}{Adult (2098)}} & & Original & \textbf{0.90} & - & - & - & - \\
\cdashline{2-8}
& \multirow{4}{*}{SynthLoop} & Info - Full & 0.76 & 0.88 & 0.64 & 0.71 & 12.28 \\
& & Info - Weakness & 0.80 & 0.89 & 0.88 & 0.81 & 16.08 \\
& & No Info - Full & 0.73 & 0.88 & 0.81 & 0.71 & 6.32 \\
& & No Info - Weakness & 0.77 & 0.88 & 0.86 & 0.51 & 9.56 \\
\cdashline{2-8}
& \multirow{4}{*}{Prompt-Refine} & Info - Full & 0.86 & \textbf{0.89} & 0.94 & \textbf{0.86} & 3.16 \\
& & Info - Weakness & 0.74 & 0.88 & \textbf{1.00} & 0.50 & 0.00 \\
& & No Info - Full & 0.81 & 0.88 & 0.91 & 0.79 & 1.76 \\
& & No Info - Weakness & 0.74 & 0.88 & 0.75 & 0.66 & 0.48 \\
\hline
\multirow{9}{*}{\rotatebox[origin=c]{90}{Thyroid (306)}} & & Original & 0.97 & - & - & - & - \\
\cdashline{2-8}
& \multirow{4}{*}{SynthLoop} & Info - Full & 0.95 & 0.98 & 0.68 & 0.78 & 0.44 \\
& & Info - Weakness & 0.89 & 0.96 & 0.54 & 0.69 & 0.52 \\
& & No Info - Full & 0.91 & 0.96 & 0.64 & 0.53 & 0.16 \\
& & No Info - Weakness & 0.97 & \textbf{0.98} & 0.72 & 0.78 & 0.24 \\
\cdashline{2-8}
& \multirow{4}{*}{Prompt-Refine} & Info - Full & \textbf{0.99} & 0.97 & \textbf{0.93} & \textbf{0.85} & 7.08 \\
& & Info - Weakness & 0.96 & 0.97 & 0.92 & 0.78 & 2.80 \\
& & No Info - Full & 0.98 & 0.97 & 0.92 & 0.82 & 14.12 \\
& & No Info - Weakness & 0.74 & 0.97 & 0.62 & 0.60 & 0.04 \\
\hline
\end{tabular}
}
    \label{tab:info_weakness}
\end{table}

\subsection{Prompt Variants}



This experiment studies the impact of prompt designs on generation and feedback. As stated in section \ref{sec:synthloop}, the generation prompt contains feature details, including possible values for each of them. A variation is to look at the impact of the presence (Info) or absence (No Info) of such information. For the feedback, the prompt tasks the LLM to analyze both strengths and weaknesses of generated data (Full) or only the weaknesses (Weakness), with the goal of focusing the feedback on the problems to correct during the generation. The default version from previous tables corresponds to "Info - Full".

Table \ref{tab:info_weakness} shows the results for a subsample of methods and datasets. The default setup with all information in both cases tends to provide the best results overall, although other prompt designs also perform well.
The patterns shown in the few-shot examples may be enough for the model to output examples similar to the real ones, even for the Thyroid case with arguably no data contamination.
We hypothesize that the additional information about the dataset may still better direct the generation of the data than by only using few-shot examples (as in EPIC).
Making the feedback list strengths along with weaknesses may also help to get better data as it may preserve desirable traits of synthetic data.

\section{Conclusion}

In this work, we proposed TAGAL, a collection of agentic, LLM training-free methods that use automatic refinement leveraging LLMs and feedback to generate tabular data. We showed that TAGAL can generate synthetic data whose quality is close to the one obtained with state-of-the-art generative models that require training while mostly outperforming other training-free approaches. We believe that this work can open the path to new methods that generate synthetic tabular data using agentic workflows.
We opt here for a feedback approach but underline that another possibility may be to use search and reinforcement learning \cite{guo2025deepseek}.
In future works, we plan to use additional datasets to deepen the comparison of TAGAL with the state-of-the-art.
Different prompt variants, for example for conditional generation of examples, can also be explored. The method's performance when very few data are available can be studied as well.


\begin{credits}
\subsubsection{\ackname} Computational resources have been provided by the supercomputing facilities of the Université catholique de Louvain (CISM/UCL) and the Consortium des Équipements de Calcul Intensif en Fédération Wallonie Bruxelles (CÉCI) funded by the Fond de la Recherche Scientifique de Belgique (F.R.S.-FNRS) under convention 2.5020.11 and by the Walloon Region.

\subsubsection{\discintname}
The authors have no competing interests to declare that are
relevant to the content of this article.
\end{credits}
%
%
%
\bibliographystyle{splncs04}
\bibliography{refs}
%
\end{document}